\documentclass[submission,copyright,creativecommons]{eptcs}
\providecommand{\event}{ICLP 2023} 

\usepackage[numbers,sort]{natbib}

\usepackage{iftex}

\ifpdf
  \usepackage{underscore}         
  \usepackage[T1]{fontenc}        
\else
  \usepackage{breakurl}           
\fi

\usepackage{amsmath}
\usepackage{amssymb}
\usepackage{graphicx}
\usepackage{multirow}

\usepackage{xcolor}
\usepackage{color}

\usepackage{adjustbox}

\usepackage[ruled,vlined,linesnumbered]{algorithm2e}

\usepackage{subcaption}

\usepackage{fontawesome5}

\DeclareMathAlphabet{\mathcal}{OMS}{cmsy}{m}{n}

\usepackage{colortbl}
\makeatletter
\def\hlinewd#1{%
  \noalign{\ifnum0=`}\fi\hrule \@height #1 \futurelet
   \reserved@a\@xhline}
\makeatother

\colorlet{DarkBlue}{blue!50!black}

 \newenvironment{graybox}[1]{%
    \begin{adjustbox}{minipage=[b]{\textwidth},margin=1ex,bgcolor=gray!13,env=center}
     \hfill{\sffamily\textbf{#1}}\\[4pt]
}{%
    \end{adjustbox}%
}

\usepackage{graphicx}

\title{{{Assessing Drivers' Situation Awareness in\\Semi-Autonomous Vehicles}}\\[6pt] \smallskip 
{\normalfont{\Large ASP based Characterisations of Driving Dynamics for Modelling Scene Interpretation and Projection}}}
\author{Jakob Suchan \quad\quad\quad\quad Jan-Patrick Osterloh
\institute{German Aerospace Center (DLR)}
\institute{Institute for Systems Engineering for Future Mobility, Oldenburg, Germany}
\email{jakob.suchan@dlr.de, jan-patrick.osterloh@dlr.de}
}
\def\titlerunning{Assessing Drivers' Situation Awareness in Semi-Autonomous Vehicles}
\def\authorrunning{Jakob Suchan and Jan-Patrick Osterloh}

\newcommand{\internalLinkColor}{darkmidnightblue}
\newcommand{\citationColor}{darkcerulean}
\newcommand{\urlColor}{\citationColor}

\definecolor{pitchblack}{cmyk}{1 1 1 1}
\definecolor{coolblack}{rgb}{0.0, 0.18, 0.39}
\definecolor{darkcerulean}{rgb}{0.03, 0.27, 0.49}
\definecolor{darkmidnightblue}{rgb}{0.0, 0.2, 0.4}

\hypersetup{
    pdftitle={Assessing Drivers' Situation Awareness in Semi-Autonomous Vehicles: ASP based Characterisations of Driving Dynamics for Modelling Scene Interpretation and Projection}, 						
    pdfauthor={Jakob Suchan, Jan-Patrick Osterloh},     					
    pdfsubject={Assessing Drivers' Situation Awareness in Semi-Autonomous Vehicles; Application of Logic Programming}, 									
    pdfkeywords={Autonomous Driving; Situation Awareness; Answer Set Programming; Declarative Dynamics; Interpretation and Projection}, 						
    unicode=false,          									
    pdftoolbar=false,        									
    pdfmenubar=true,        								
    pdffitwindow=false,     									
    pdfstartview={FitH},    								
    pdfnewwindow=true,      								
    bookmarks=true,         								
    colorlinks=true,       									
    linkcolor=\internalLinkColor,          							
    citecolor=\citationColor,        							
    filecolor=\internalLinkColor,      									
    urlcolor=\urlColor          						
}

\newenvironment{prolog-small}{
    \vspace{-5pt}
    \VerbatimEnvironment
    \noindent\begin{minipage}{\linewidth}
    \begin{minted}[
    fontsize=\tiny,
    bgcolor=blue!10
    ]{prolog}}
{\end{minted}\end{minipage}\vspace{-5pt}}
 
\newenvironment{result-small}{ 
    \vspace{-5pt}
    \VerbatimEnvironment
    \noindent\begin{minipage}{\linewidth}
    \begin{minted}[
    fontsize=\tiny,
    bgcolor=green!25
    ]{prolog}}
{\end{minted}\end{minipage}\vspace{-5pt}}

\begin{document}

\maketitle

\begin{abstract}

Semi-autonomous driving, as it is already available today and will eventually become even more accessible, implies the need for driver and automation system to reliably work together in order to ensure safe driving. A particular challenge in this endeavour are situations in which the vehicle's automation is no longer able to drive and is thus requesting the human to take over. In these situations the driver has to quickly build awareness for the traffic situation to be able to take over control and safely drive the car.
Within this context we present a software and hardware framework to asses how aware the driver is about the situation and to provide human-centred assistance to help in building situation awareness. 
The framework is developed as a modular system within the Robot Operating System (ROS) with modules for sensing the environment and the driver state, modelling the driver's situation awareness, and for guiding the driver's attention using specialized Human Machine Interfaces (HMIs).

\smallskip

A particular focus of this paper is on an Answer Set Programming (ASP) based approach for modelling and reasoning about the driver's interpretation and projection of the scene. This is based on scene data, as well as eye-tracking data reflecting the scene elements observed by the driver. 
We present the overall application and discuss the role of semantic reasoning and modelling cognitive functions based on logic programming in such applications. Furthermore we present the ASP approach for interpretation and projection of the driver's situation awareness and its integration within the overall system in the context of a real-world use-case in simulated as well as in real driving.

\end{abstract}


\section{Introduction}

With the rise of automated and semi-automated driving, a range of new challenges have moved into sight of developers. In particular semi-autonomous vehicles pose important questions when it comes to the interplay between driver and vehicle, since for these systems it is critical that the cooperation of the human and the automation system is seamless.  
To ensure this, the automation system needs to be designed with the needs of the human in mind and ideally have the functionality to asses the mental state of the driver and the driver's interaction with the system. Such functionality is especially important in situations when the driver and the automation have to function together, in order to safely operate the vehicle. 
A particular challenge in this context, and of importance with respect to this paper, is the question of how to enable the driver to quickly take over control from the automation system in situations, in which the automation cannot ensure safe driving, e.g., in unforeseen situations, or in situations outside of the Operational Design Domain (ODD) of the automation.

\medskip

\noindent The SituWare project presented in this application paper aims at developing a system to provide the driver with assistance in such situations, by modelling the driver's awareness of the situation and guide their attention to critical elements of the situation possibly missed by the driver. The focus of this paper is on highlighting the application of declarative logic programming methods as a means to model the driver's mental processing of the situation and demonstrate its integration within a large-scale modular assistive system. 
The work presented in this paper is driven by considerations in the fields of human-centred design, cognition, and formal semantic reasoning.


\medskip

\noindent\textbf{Human-centred design for (semi-)autonomous vehicles.} \quad
With the above challenges in mind human factors play a central role in the development of novel mobility systems and autonomous vehicles \citep{kyriakidis2019-survey-human-factors} and development of novel systems is often accompanied and driven by empirical research investigating how humans interact with these systems, either for external communication with vulnerable road users, e.g. \citep{zimmermann2017first,rasouli2020}, and for internal communication e.g. in handover scenarios \citep{van2017priming,morgan2018,hsiung2022} or for acceptance and user experience \citep{roedel2014}.

\medskip

\noindent\textbf{Cognitive abilities for driving assistance systems.} \quad
Research in the area of cognitive systems and cognitive modelling is concerned with developing methods and tools that reflect cognitive abilities. 
Most directly connected to the topic of this paper and the development of SituSYS is the research on Situation Awareness and its modelling in computational systems \citep{endsley95}. In the context of driving assistance systems research on Situation Awareness has mostly been concerned with modelling e.g. by \citep{rehman2019,krems2009,iccm2017}.

\medskip

\noindent\textbf{Semantic reasoning about dynamic situations.} \quad
A key factor in developing such capabilities is the ability to abstractly represent the driving environment and the traffic dynamics within it, to interpret and reason about them on a semantic level. 
In this context logic programming and Answer Set Programming (ASP) \citep{Gebser2012-ASP,Gebser2014-Clingo,Gebser2016-clingo5} in particular has evolved as a powerful tool for semantic reasoning about dynamic scenes. For instance, ASP has been used in the area of stream reasoning \citep{DBLP:conf/ecai/EiterK20} as a general tool for reasoning about large scale dynamic data, e.g, in the driving domain \citep{Le-Phuoc-Eiter-Le-Tuan-2021}, for decision-making \citep{DBLP:conf/iclp/KothawadeK0WG21}, or for recognition and reasoning about events \citep{tsilionis-koutroumanis-nikitopoulos-doulkeridis-artikis-2019}.  
Furthermore, \cite{Suchan18} has developed a general method for visual abduction based on ASP, and applied it in the area of online semantic sense-making with commonsense knowledge in the context of safety-critical situations in driving \citep{DBLP:journals/ai/SuchanBV21,Suchan-IJCAI2019}.

\medskip

Within this paper we build on these works to develop an ASP-based approach for assessing the driver's situation awareness by characterising driving dynamics for modelling and reasoning about the driver's interpretation and projection of the scene dynamics.

\medskip

\section{SituWare: An Online System for Assessing Situational Awareness}
\label{sec:situware}

The general focus of the SituWare project is f build and evaluate the technological basis for developing a usable software and hardware system (\emph{SituSYS}), targeted at the reliable detection, interpretation, and consideration of the driver's situational awareness. In this section we discuss the relevance of situation awareness in (semi-)autonomous driving and provide an overview of the full system and its components.

\subsection{Situation Awareness in (Semi-)Autonomous Driving}

Today, in the automotive domain and within the framework of the SAE J3016 automation level system, it is generally assumed that the driver observes the environment and is thus immediately available as a fallback level (SAE level 2 - partial automation), can react to a request to intervene (SAE level 3 - conditional automation) or can take over the vehicle control outside the operational area of the automation (SAE level 4 - high automation). Previous studies on handover scenarios, especially in the context of conditional automation, show that a considerable amount of time of at least 7 to 10 seconds must be reserved for the driver to safely take over the driving task. This time is needed by the driver(s) to gain an accurate situation awareness so that a safe handover is accomplished. Current approaches for handover scenarios from the system to the driver consider the current driver's state only peripherally and very roughly. However, the potential of such recognition and interpretation is very large. Adaptive interaction sequences can be used to optimally return the driver to the driving task. In addition, a more detailed picture of the driver's state also allows for the adaptation of the driving behavior of the automation system, so that larger time reserves can be kept available for a highly distracted driver, while time reserves can be reduced for an attentive driver without reducing safety. 
One of the most prominent models for Situation Awareness is the one developed by Mica Endsley \citep{endsley95}.
The idea behind this model is that Situation Awareness is built in three Stages, namely the \emph{Perception}, \emph{Interpretation}, and \emph{Projection} stage, also known as Level 1 to Level 3 Situation Awareness ({\sffamily\color{blue!70!black}\textbf{L1}} - {\sffamily\color{blue!70!black}\textbf{L3}}): 

\medskip
\noindent {\sffamily\color{blue!70!black}\textbf{L1.}} \quad At Level 1, the data and elements of the environment that build the current situation are perceived. The main factor that influences Level 1 Situation Awareness is the focus of attention that is apparent in the situation, which in turn is guided by the goals and objectives of the user, as well as their expectations. Previous Experience and Training can influence the attention process and thus the ability of the user to perceive the environment correctly. Drivers need to perceive the cars around their own car, signs along the road, the road curvature and condition, and lots of other things. 

\medskip
\noindent {\sffamily\color{blue!70!black}\textbf{L2.}} \quad Level 2 Situation Awareness builds upon Level 1 through interpretation and comprehension of the perceived elements. At Level 2, a holistic picture of the environment including the significance of objects and events is formed. This interpretation is influenced by the goals of the user, previous experience as well as the workload and stress. In the automotive example, drivers need to understand that proximity of other cars might indicate a risk, that there are speed limits to keep, or the importance of the other signs on the road.

\medskip
\noindent {\sffamily\color{blue!70!black}\textbf{L3.}} \quad At Level 3, the user builds a projection of the current situation into the future, based on the comprehension built at Level 2. Level 3 Situation Awareness is mainly influenced by the previous experience and training of the user, since this builds his knowledge on how the entities in the environment may act in the near future. For drivers, this could mean to predict the future speed and positions of other cars as well as possible actions the other drivers might take (like changing lanes, or breaking) as well as upcoming speed limits and the risk associated with the road conditions ahead.

\begin{figure}[t]
\centering
\includegraphics[width=0.75\textwidth]{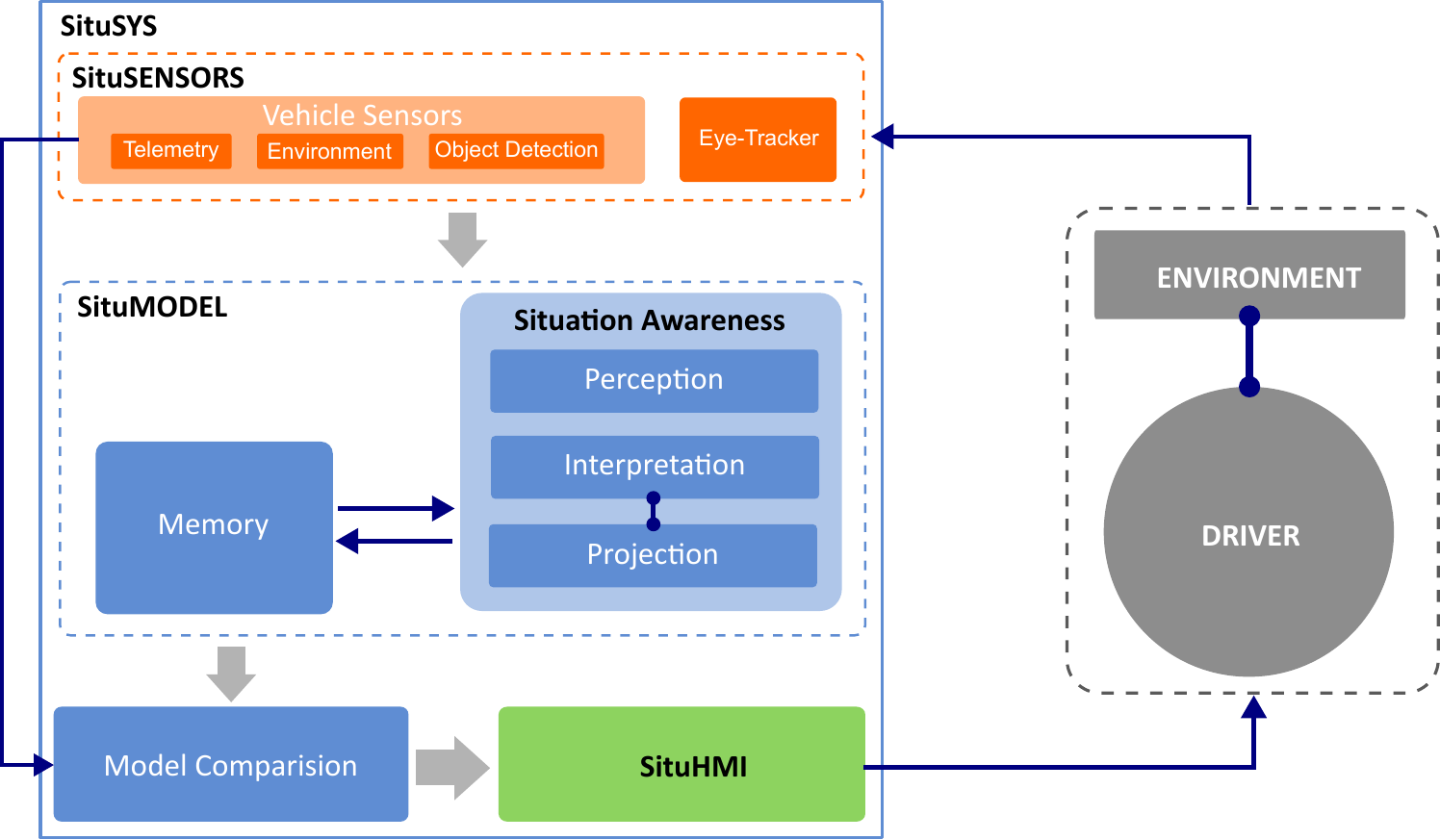}
\caption{\textbf{SituSYS}: Conceptual Overview}
\label{fig:concept-overview}
\end{figure}

\subsection{A Modular System implemented in ROS}

Objective of the SituWare project is to build \emph{SituSYS}, a system that predicts the situation awareness of a driver, calculates possible deviations from an optimal situation awareness and finally uses specialized interaction techniques to improve the driver's situation awareness. Figure \ref{fig:concept-overview} shows a conceptual overview of the architecture of \emph{SituSYS}: \emph{SituSYS} consists of three parts, the sensor layer \emph{SituSENSORS} which measures the driver and environment, the model layer \emph{SituMODEL} which calculates the situation awareness, and \emph{SituHMI} for the interaction. 
Starting at the sensor-layer \emph{SituSENSORS}, multiple \emph{Vehicle Sensors} are used to sense the environment and vehicle state, i.e. surrounding objects, signs, current vehicle speed and automation state. In addition to the vehicle sensors, an eye-tracker is used to detect the gaze of the driver. The calculated gaze vector is then used in the \emph{Perception}-Model within \emph{SituMODEL} to predict which elements in the environment and in the car has been looked at, i.e. other cars, street signs, or any cockpit elements. Based on this information, the objects that have been perceived are then written into the \emph{Memory}-Model. The \emph{Memory}-Model implements retrieval and forgetting processes, and can be accessed by the other \emph{SituMODEL} components.  The \emph{Interpretation}- and \emph{Projection}-Models are for the calculation of the situation interpretation and the projection of the future state  (A detailed description can be found in Section \ref{sec:iup-model}). 
The outputs of the \emph{Perception}-, \emph{Interpretation}-, and \emph{Projection}-Models are then used by the \emph{Model Comparison} to calculate the overall situation awareness (Level 1, Level 2 and Level 3 in Endsley's Model), by comparing the content in the memory, that has been produced by the components of the \emph{SituMODEL}, with the data gathered by the vehicle sensors. Although the vehicle sensors have an error probability associated with them, we assume in this case that this is the \emph{ground truth} for \emph{SituSYS}. The \emph{Model Comparison} then weights the differences between the memory and the ground truth and calculates a list of diverging elements sorted by priority. This list is then used by \emph{SituHMI} to direct the focus of attention to the most important element. In SituWare we tested different interaction methods for that purpose \citep{interactionPaper}. 

\medskip

To organize and coordinate the implementation of the different components of \emph{SituSYS}, the \emph{ROS} framework has been used. Each component of \emph{SituSYS} has been implemented as separate node, allowing a modularization of the system. \emph{ROS messages} describing the data exchanged between the components are used to provide a standardised interface between the different modules.  
The use of the ROS framework also facilitates the use of different sources for the input, i.e. we connected \emph{SituSYS} to two different driving simulators as well as to a real car, by implementing a dedicated node that collects the data from the simulator or car and then sends them in the defined messages to \emph{SituSYS}, and receives the messages from \emph{SituHMI} to implement the selected interaction method in the vehicle/simulator.

\begin{figure}
\centering
\subcaptionbox*{a) Simulated scene of construction site after takeover}
{\includegraphics[height=4.0cm]{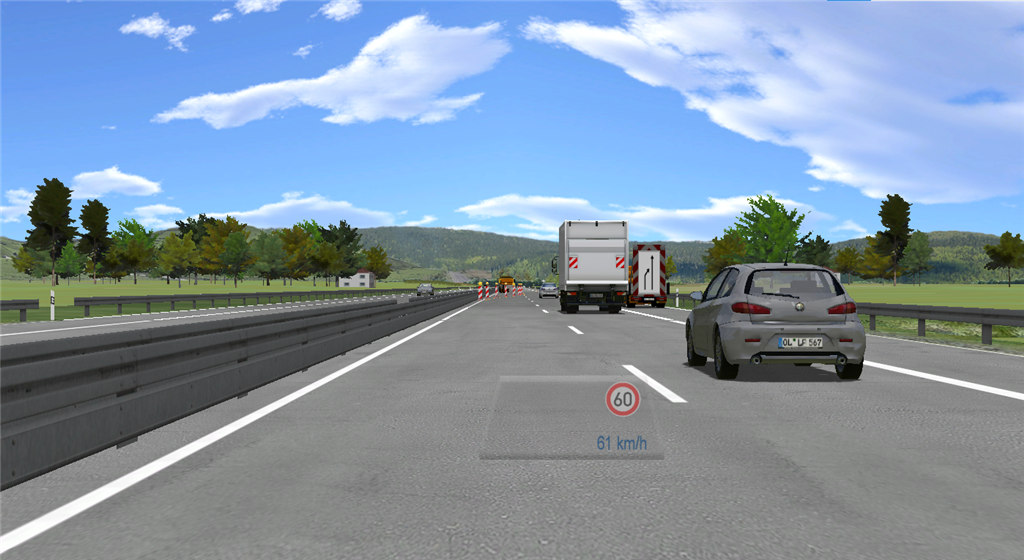}}
\quad
\subcaptionbox*{b) Driving Simulator}{\includegraphics[height=4.0cm]{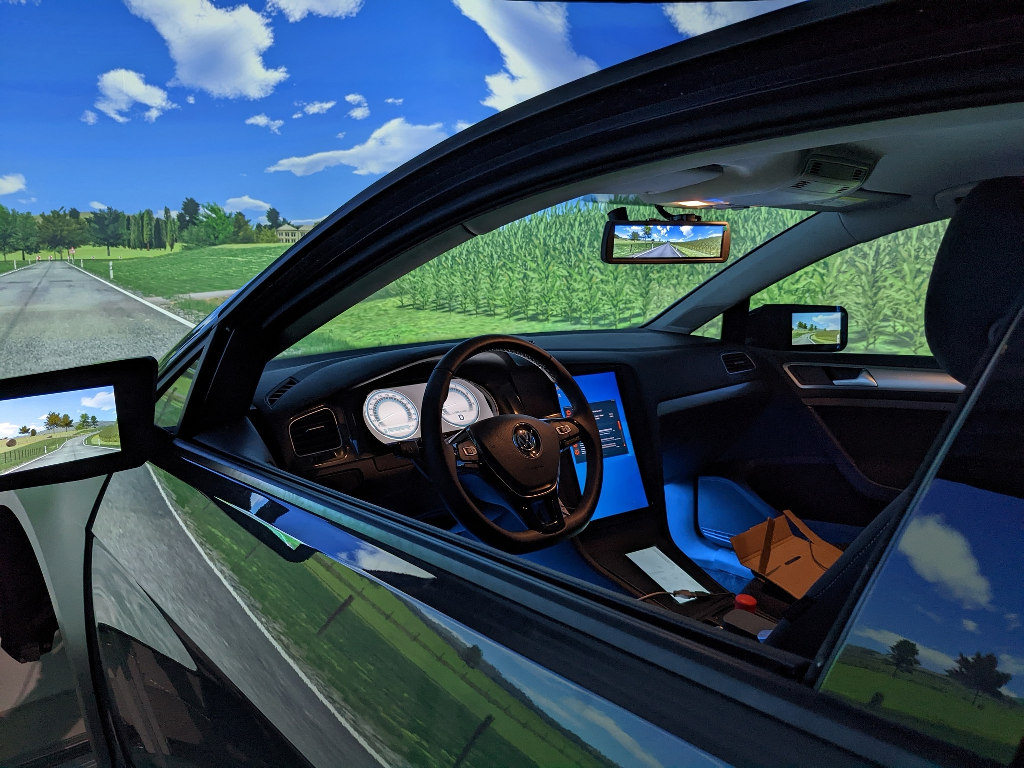}}

\caption{\textbf{Simulation Environment:} The simulated scene and the simulator setup.}
\label{fig:simulation}
\end{figure}

\subsection{Technical Setup and Data}
\label{sec:data}

The SituWare project integrates SituSYS within two different simulators and one real semi-autonomous vehicle and provides the hardware basis for sensing the ego vehicle and the environment together with mobile tracking of the driver's gaze. 
Figure \ref{fig:simulation} shows the simulated scenario within one of the two simulators used within the SituWare project.

\medskip

\noindent \textbf{{\sffamily$\blacktriangleright$}} \hspace{0.2cm} {\sffamily\textbf{Scene Data}} \quad
The vehicle sensors implemented in the specific platform are used to sense the state of the ego vehicle and to sense the environment including the vehicles within it. This information is published as \emph{ROS messages} and used by the modules of \emph{SituSYS}. The messages contain data on the ego vehicle, the automation system, and other vehicles in the scene as listed in table \ref{tbl:scene-data}.

\medskip

\noindent \textbf{{\sffamily$\blacktriangleright$}} \hspace{0.2cm} {\sffamily\textbf{Eye-Tracking Data}} \quad
The driver's perception is sensed using mobile eye-tracking that provides gaze coordinates and fixations within an ego view of the scene, as captured by the camera of the eye tracker. These coordinates are translated to the 3-dimensional space of the scene, which allows mapping of gaze data to scene elements. Within \emph{SituSYS} the gaze data is provided as a 3D vector. The modular design of the system abstracts from the actual eye-tracking system, however, in the case study presented in this paper the PupilLabs mobile eye-tracker is used, which is equipped with a eye camera capturing the gaze with up to $120 Hz$  and providing an accuracy of $0.60^{\circ}$.

{
\begin{table}[t]
\begin{center}
\footnotesize
\begin{tabular}{>{\columncolor[gray]{0.92}}l l p{6.8cm}}
\hlinewd{1pt}
{\sffamily\color{blue!70!black}\textbf{Attribute}} & \textbf{Type} & \textbf{Description} \\

\hline
\multicolumn{3}{c}{Ego Vehicle}\\\hline

ID & \emph{string} & The $ID$ of the ego vehicle.\\
type & \emph{string} & The $type$ of the ego vehicle.\\
position & \emph{\{float, float, float\}} & The $position$ of the ego vehicle as 3D vector.\\
orientation & \emph{\{float, float, float\}} & The $orientation$ of the ego vehicle as 3D vector.\\
velocities & \emph{\{float, float, float\}} & The $velocity$ of the ego vehicle as 3D vector.\\
indicator\_left & \emph{bool} & Truth value whether the $left\ indicator$ is active.\\
indicator\_right & \emph{bool} & Truth value whether the $right\ indicator$ is active.\\
acceleration & \emph{float} & The $acceleration$ of the ego vehicle.\\
current\_speed\_limit & \emph{int} & The $current\ speed\ limit$ holding for the ego vehicle.\\
current\_lane & \emph{float} & The $current\ lane$ the ego vehicle is on.\\

\hline
\multicolumn{3}{c}{Ego Automation}\\\hline

takeover\_request & \emph{bool} & Truth value whether the $takeover\ request$ is active.\\
time\_until\_odd\_boundary & \emph{float} & $Time$ until the driver has to take over.\\
criticality\_level & \emph{int} & $Criticality\ level$ of the takeover request.\\
takeover\_reason & \emph{string} & $Reason$ for the takeover request.\\
ego\_automation\_state & \emph{bool} &Truth value whether the $automation$ is active.\\

\hline
\multicolumn{3}{c}{Other Vehicles}\\\hline

ID & \emph{string} & The $ID$ of the traffic vehicle.\\
type & \emph{string} & The $type$ of the traffic vehicle.\\
position & \emph{\{float, float, float\}} & The $position$ of the traffic vehicle as 3D vector.\\
orientation & \emph{\{float, float, float\}} & The $orientation$ of the traffic vehicle as 3D vector.\\
velocities & \emph{\{float, float, float\}} & The $velocity$ of the traffic vehicle as 3D vector.\\
acceleration & \emph{\{float, float, float\}} & The $acceleration$ of the traffic vehicle as 3D vector.\\
dimension & \emph{\{float, float, float\}} & The $dimension$ of the traffic vehicle as 3D vector.\\
lane & \emph{int} &  The $lane$ the traffic vehicle is on.\\

fixation\_probability & \emph{float} & The $probability$ that the driver fixated the vehicle.\\
fixation\_time & \emph{int} & The $duration$ of the driver's fixation on the vehicle.\\

\hlinewd{1pt}
\end{tabular}
\caption{\textbf{Scene Data:} Relevant data-points from \emph{SituSYS}.}
\label{tbl:scene-data}
\end{center}
\end{table}%
}

\section{Assessing the Driver's Situation Awareness}
\label{sec:iup}

Within \emph{SituSYS} situation awareness is modelled by the perception module and the interpretation \& projection module. The perception module implements 1st level Situation Awareness using scene data from the \emph{SituSENSORS} together with eye-tracking data to calculate fixation probabilities for each scene object. 

\vspace{-10pt}

\begin{graybox}{{\color{DarkBlue} {\faEye}} $~$ Perception}
Level 1 Situation Awareness --  \emph{perception of the scene}\footnotemark  -- is estimated based on the gaze data of the driver which is recorded by \emph{SituSENSORS}. From this information a probability is computed for each scene object giving the likelihood that the driver has fixated the object together with a fixation time providing the duration of the fixation on the object.
These measures are then published within the \emph{ROS message} as an input for the interpretation \& projection module, as well as the model comparison. 
\end{graybox}

\footnotetext[1]{Technical details on the used method for calculating the probabilities are out of the scope of this paper. For the examples of this paper we consider the perception module to be a black-box system providing the information if an object was perceived by the driver.}

\noindent 2nd and 3rd level Situation Awareness are implemented together in the interpretation \& projection module. These consist of:

\begin{itemize}

\item Level 2 Situation Awareness -- \emph{interpretation of the scene} -- models the awareness of the driver regarding scene elements in the current situation;

\item Level 3 Situation Awareness -- \emph{projection of scene dynamics} -- models the expectations of the driver how the scene will evolve, i.e., the scene dynamics with respect to the task of the driver.

\end{itemize}

\noindent The implementation of these levels is based on semantic characterisations of the dynamics of the driving domain, which are declaratively defined within answer set programming (ASP) \citep{Gebser2012-ASP,Gebser2014-Clingo,Gebser2016-clingo5}. 
In particular, the interpretation \& projection module of \emph{SituSYS} consist of an online Python process maintaining a representation of the scene and uses an integrated ASP solver to generate interpretation and projection models based on the scene data from the \emph{SituSENSORS} and the perception data from the perception module.
In the following we provide the formal characterisation of the driving domain, and describe the interpretation and projection process in detail.

\subsection{The Driving Domain}
\label{sec:driving-domain}

The domain is characterised by $\Sigma$$<$$\mathcal{O}$, $\mathcal{E}$, $\mathcal{R}$, $\mathcal{T}$, $\Phi$, $\Theta$$>$, which is used to formalise the driver's representation of the scene dynamics.
In particular, the driver's belief state is represented on the one hand by the static and dynamic properties of the scene and the elements within it given by the domain objects, the basic entities representing these objects, and the spatial and temporal aspects of the scene $<\mathcal{O}$, $\mathcal{E}$, $\mathcal{R}$, $\mathcal{T}>$. On the other hand it is represented by the high-level event dynamics of perceived events and possible future events given by  $<\Phi$, $\Theta>$.

\medskip
\noindent \textbf{{\sffamily$\blacktriangleright$}} \hspace{0.2cm} {\sffamily\textbf{Domain Objects ($\mathcal{O}$) and Spatial Entities ($\mathcal{E}$).}} \quad
The scene consists of different scene elements, in particular we are considering the ego vehicle and other vehicles in the scene, as well as the lanes on the road and gaps between vehicles.

\medskip

\noindent\emph{Vehicles.}\quad We distinguish between the ego vehicle and other vehicles in the scene, constituting the traffic. We use the following objects for representing vehicles:

\begin{itemize}

\item The \emph{ego vehicle}:  $\mathcal{O}_{ego} = ego$;  
\item other vehicles in the traffic:  $\mathcal{O}_{trf} =\{trf_1, ..., trf_n \}$.

\end{itemize}

\noindent These elements are geometrically represented as spatial entities $\mathcal{E} = \{\varepsilon_{1}, ..., \varepsilon_{n}\}$ within the 3-dimensional scene space. Additionally they are assigned dynamic and static attributes as obtained from the egos driving system (for the attributes of the ego vehicle) and the ego vehicle sensors (for estimated attributes of other vehicles) as detailed in section \ref{sec:data}.   

\medskip

\noindent\emph{Lanes.}\quad These are based on the OpenDRIVE standard \citep{OpenDrive}
, in which lanes are numbered with positive and negative numbers and 0 represents the middle of the road. Lanes are adjacent when the lane $ids$ are consecutive. 

\begin{itemize}

\item Lanes:  $\mathcal{O}_{lanes} = \{lane_1, ..., lane_n \} \cup \{lane_{-1}, ..., lane_{-n} \}$.

\end{itemize}

\noindent We define the adjacency of lanes  $lane_i, lane_j \in \mathcal{O}_{lanes}$ on the road using the predicate $adjacent(lane_i, lane_j)$.

\medskip

\noindent\emph{Gaps.}\quad
Of particular interest for the task at hand are gaps between vehicles in the scene. Therefore we introduce objects representing these gaps.

\begin{itemize}

\item Gaps:   $\mathcal{O}_{gaps} = \{gap_{{trf_i}, {trf_j}}, ..., gap_{{trf_p}, {trf_q}}\}$.

\end{itemize}

\noindent Gaps are declaratively defined based on the vehicles in the scene using the predicate $gap(trf_i, trf_j)$, where $trf_i, trf_j \in \mathcal{O}_{trf}$. Additionally we define the predicate $gap\_size(gap_{{trf_i}, {trf_j}}, size)$, where $gap_{{trf_i}, {trf_j}} \in \mathcal{O}_{gaps}$ and $size$ is a number representing the distance between $trf_i$ and $trf_j$.

{
\begin{table}[t]
\begin{center}
\footnotesize
\begin{tabular}{>{\columncolor[gray]{0.92}}l p{6.5cm}}
\hlinewd{1pt}
{\sffamily\color{blue!70!black}\textbf{Events}} & \textbf{Description} \\\hline

$change\_lane(Entity, Lane_1, Lane_2, Location)$ & An Entity changing the lane from $Lane_1$ to $Lane_2$ to a specific $Location$.\\

audio\_signal\_start & The audio signal indicating that the driver has to take over started.\\

audio\_signal\_end & The audio signal indicating that the driver has to take over ended.\\

takeover\_manual & The driver takes over control of the vehicle.\\

takeover\_automation & The automation takes over control of the vehicle.\\

\hlinewd{1pt}
\end{tabular}

\medskip

\begin{tabular}{>{\columncolor[gray]{0.92}}l p{9.65cm}}
\hlinewd{1pt}

{\sffamily\color{blue!70!black}\textbf{Fluents}} & \textbf{Description} \\\hline

curr\_task(Task)  & The current task, the driver has to perform.\\

automation &  If the driving automation of the vehicle is on or not.\\

audio\_signal &  If the audio signal indicating that the driver has to take over control of the vehicle is on or not.\\

on\_lane(Entity, Lane) & The lane a scene entity is driving on.\\

\hlinewd{1pt}
\end{tabular}
\caption{{ {\bf Event Dynamics:} Exemplary events and fluents applicable in take-over situations of the use-case described in section \ref{sec:iup-model}.}}
\label{tbl:event-dynamics}
\end{center}
\end{table}%
}

\medskip
\noindent \textbf{{\sffamily$\blacktriangleright$}} $~~$ {\sffamily\textbf{Spatial Configuration ($\mathcal{R}$).}} \quad
The spatial arrangement of the entities in the scene is represented from an egocentric perspective in the context of the road, i.e., representing whether vehicles are ahead or behind the ego vehicle, on the same lane, or on the lane to the left or right of the ego vehicle, and how many other vehicles are between a vehicle and the ego vehicle. 
To this end we define the following relations in $\mathcal{R}$ holding between the ego vehicle and the other scene elements.

\begin{itemize}

\item \emph{Relative longitudinal direction}:  $R_{rel\_long} = \{ahead, behind, overlapping\}$, representing the direction of an object on the longitudinal axis based on the road layout. 

\item \emph{Relative lane}: $R_{rel\_lane} = \{same, left, right\}$, representing the lane an object is on, relative to the lane the ego vehicle is on.

\item \emph{Relative ordering}:  $R_{rel\_order} = n +1$, where $n$ is the number of other vehicles between the vehicle and the ego vehicle, representing the position of an object with respect to the ego vehicle and the other vehicles on the road.

\end{itemize}

\medskip
\noindent \textbf{{\sffamily$\blacktriangleright$}} $~~$ {\sffamily\textbf{Time ($\mathcal{T}$).}} \quad
We represent time using time points $\mathcal{T}= \{t_{1}, ..., t_{n}\}$. These are used to denote that dynamic object properties and spatial relations between basic entities representing scene objects hold at a certain time, as well as to describe temporal aspects of event dynamics. 

\medskip
\noindent \textbf{{\sffamily$\blacktriangleright$}} $~~$ {\sffamily\textbf{Driving Event Dynamics ({$<${$\Phi$}}, {{$\Theta$}$>$).}}} \quad
We use the event calculus notation to define the event dynamics in the driving domain. Towards this we define  \textbf{fluents} ${\Phi} = \{\phi_1, ... , \phi_n\}$ and \textbf{events} ${\Theta} = \{\theta_1, ... , \theta_n\}$ to characterise dynamic properties of the scene objects and high-level events (e.g., Table \ref{tbl:event-dynamics}).  
We use the axioms ${occurs\_at}(\theta, t)$ denoting that an event occurred at time $t$ and ${holds\_at}(\phi, v, t)$ denoting that $v$ holds for a fluent $\phi$ at time $t$.

\begin{figure}[t]
\centering
\includegraphics[width=1.0\textwidth]{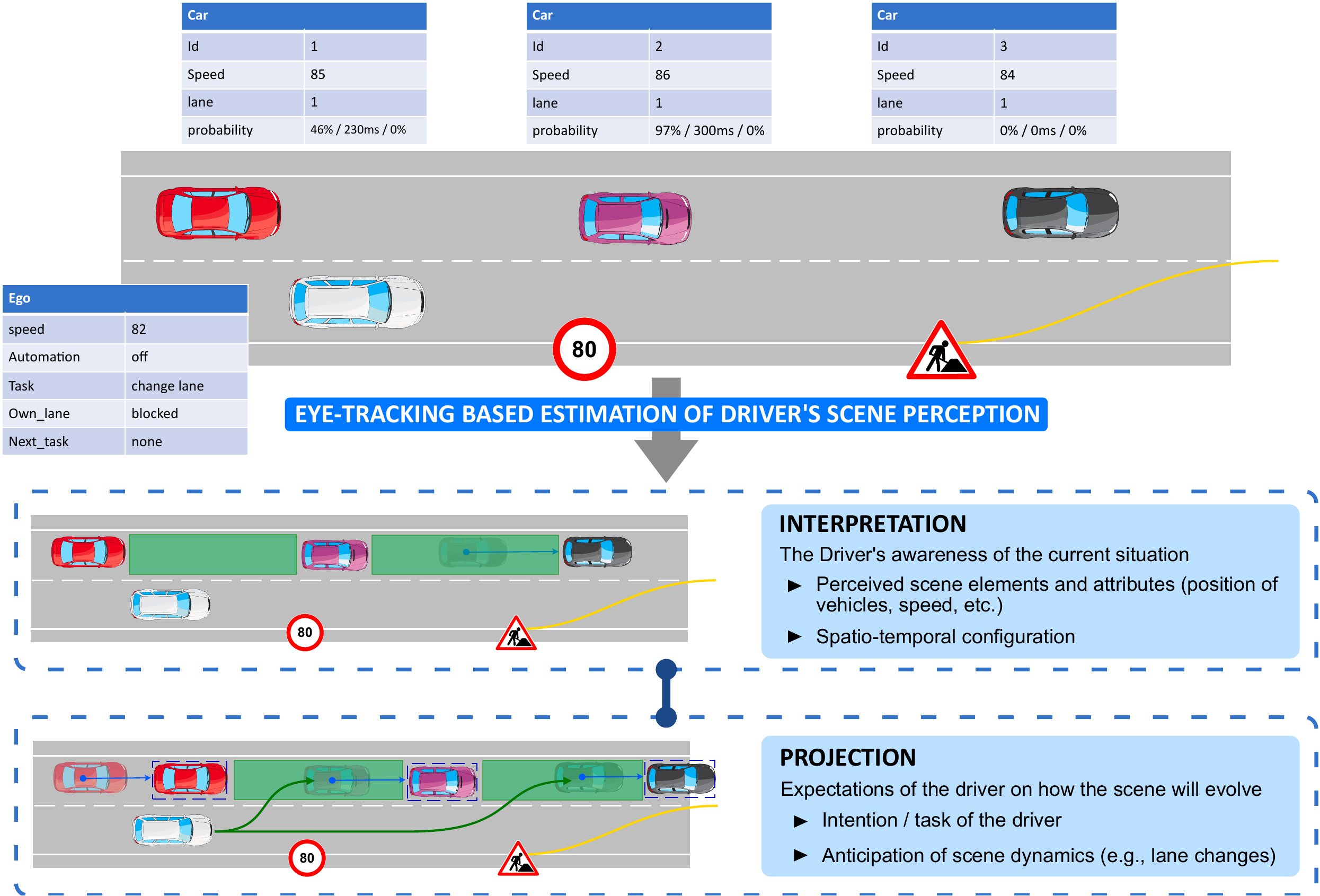}

\caption{\textbf{Application Example:} Interpretation \& Projection}
\label{fig:SituWare-example}
\end{figure}

\subsection{Reasoning about Situation Awareness Level 2 \& 3: Interpretation and Projection}\label{sec:iup-model}

The interpretation and projection module is based on declarative characterisations of driving dynamics (as defined in section 
\ref{sec:driving-domain}
) implemented in ASP, using Event Calculus \citep{eventCalculus1989} for reasoning about events in the scene. In particular, we are building on the Event Calculus (EC) as formalised in \citep{Ma2014-ASP-based-Epistemic-Event-Calculus,Miller2013-Epistemic-Event-Calculus}.
The module is implemented as a hybrid system, in which an online Python process maintains a representation of the mental belief state of the driver and uses an integrated ASP solver for generating the interpretation and the projection model.
For this, the module uses the scene data $\mathcal{S}$ containing the ego vehicle data and the environmental data including the gaze data of the driver, together with the characterisations of the driving domain in $\Sigma$.

\begin{graybox}{{\color{DarkBlue} {}} $~$ Application: The Case of Take-Over Situations in a Construction Environment}
As a use case we have applied \emph{SituSYS} in the context of a case study conducted in the driving simulator.
The case study implemented a situation in which a driver had to take over the control of a highly automated vehicle at a section of a highway, where the current lane ended because of a construction site and the vehicle had to change lanes. The driver got notified about the upcoming take-over and had a certain time window to get familiar with the situation and take over control to manually perform the lane change. 
\end{graybox}

\noindent As a test case the overall system is applied in the context of the above case study, in which the driver has to take over control from the automation to safely drive the vehicle.
Fig. \ref{fig:SituWare-example} shows an exemplary situation in the context of the case study and depicts the interpretation and the projection step.

\medskip 

\noindent \textbf{Computational Steps for Situation Awareness Level 2 \& 3.} \quad 
The overall process for interpretation and projection consists of the following steps ({\sffamily\color{blue!70!black}\textbf{S1-S3}}, also refer to Alg. \ref{alg:interpretation}) performed at each time point:

\medskip

\noindent{\sffamily\color{blue!70!black}\textbf{S1.}} $~~$ {\sffamily\textbf{Update Scene Elements and Predict Current State.} }\quad
Update the driver's mental belief state ($\mathcal{MBS}$) with the scene elements for which the fixation probability is above the fixation threshold, and predict the current position of all scene elements that are part of the driver's mental belief state, using a Kalman Filter base motion model, assuming constant velocity.

\medskip

\noindent{\sffamily\color{blue!70!black}\textbf{S2.}} $~~$ {\sffamily\textbf{Generate the Problem Specification.}}\quad
Generate an ASP problem statement, containing the scene elements that are part of the driver's mental belief state, and the characterisations of the driving domain in $\Sigma$.

\medskip

\noindent{\sffamily\color{blue!70!black}\textbf{S3.}} $~~$ {\sffamily\textbf{Integrated ASP Solving.}}\quad
Generate the interpretation model ($\mathcal{IM}$) and the projection model ($\mathcal{PM}$) by solving the generated problem specification using the Clingo solver integrated within the Python process.

\medskip

\noindent Within this process the interpretation and the projection are implemented as follows:

\medskip
\noindent \textbf{{\sffamily$\blacktriangleright$}} $~~$ {\sffamily\textbf{Interpretation.}} \quad
The interpretation level of situation awareness is modelled as an extrapolated representation of the scene, consisting of the perceived scene elements, together with the arrangement of these elements and the events happening in the scene.
When the probability that the driver has fixated a particular object exceeds a certain threshold we register it within the scene representation and use a Kalman Filter based motion model assuming constant velocity to maintain a representation of the object while the driver is not fixating on it. When the driver is fixating the object again the estimated position is updated with the sensed one.
In this way we estimate the driver's mental belief state about the movement of scene elements.

\medskip

\noindent This estimated mental representation of the scene is then used together with the characterisations of the driving domain in $\Sigma$ to generate the interpretation model ($\mathcal{IM}$). To this end we declaratively model scene artefacts, spatial configuration, and events. 

\medskip 

\noindent \textbf{Scene Artefacts.}\quad 
These are elements of the scene that are indirectly obtained from the sensed objects. For instance, gaps between vehicles the driver is aware of, are declaratively defined using the following rule, stating that there is a gap between two entities if they are on the same lane and there is no other entity between these two entities.

\smallskip

\noindent\includegraphics[width=1.0\textwidth]{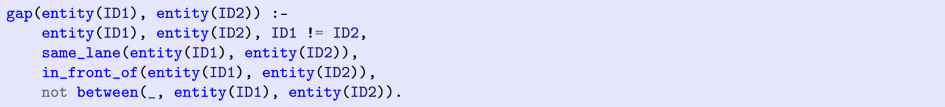}


\medskip 

\noindent \textbf{Spatial Configuration.}\quad 
The relational spatial structure holding between the scene elements the driver is aware of, is represented using the spatial relations defined in $\mathcal{R}$.

\begin{figure}[t]
\centering
\noindent\begin{minipage}{0.6\columnwidth}

{

\SetNlSty{texttt}{\color{blue}}{\quad}
\IncMargin{1.0em}

\begin{algorithm}[H]
\scriptsize

\KwData{
Scene data ({$\mathcal{S}$}), and the characterisation of the driving domain $\Sigma$ 
	}
	  \medskip

\KwResult{
Interpretation Model ({$\mathcal{IM}$}), Projection Model ({$\mathcal{PM}$})\hfill 
}

\medskip

$\mathcal{MBS} \leftarrow \varnothing$

\smallskip

\For{$t \in T$}
{

	\For{$object \in \mathcal{MBS}$}
	{
		$position_{object} \leftarrow kalman\_predict(object)$
	}
	
	\smallskip
	
	\For{$object \in \mathcal{S} \wedge not\ object \in \mathcal{MBS}$}
	{
		\If{$fixation\_probability_{object} > fixation\_threshold$}
		{
			$\mathcal{MBS} \leftarrow \mathcal{MBS} \cup object$
		}
	}
	
        \smallskip
 
	\For{$object \in \mathcal{S} \wedge object \in \mathcal{MBS}$}
	{
		\If{$fixation\_probability_{object} > fixation\_threshold$}
		{
			$position_{object} \leftarrow kalman\_update(object)$
		}
	}
	
        \smallskip
	
	$<\mathcal{IM}, \mathcal{PM}> \leftarrow ASP\_solve(\mathcal{MBS}, \Sigma)$

}

\Return{$<\mathcal{IM}, \mathcal{PM}>$}

\caption{$~~${\color{blue!80!black}{Interpret\_and\_Project}($\mathcal{S}$, $\Sigma$)}
\label{alg:interpretation}
}
\end{algorithm}
}

\end{minipage}
\end{figure}

\medskip 

\noindent \textbf{Events.}\quad 
We use the event calculus to detect driving events based on the driver's mental belief state of the scene.
Towards this we define fluents representing dynamic scene properties, e.g, the fluent $on\_lane(entity(ID), lane(Lane))$ denotes that an entity is on a particular lane, or the fluent $curr\_task(Task)$ denotes the current task of the driver.

\smallskip

\noindent\includegraphics[width=1.0\textwidth]{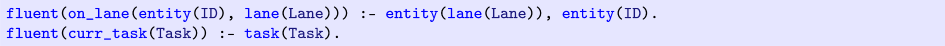}


\noindent Additionally, we define events changing these scene properties. 
For instance the following definition of the event $takeover\_manual$ states that the event occurs if the fluent $automation$ is \emph{true} and the sensed state of the automation is \emph{false}.
Further, it states that the event initiates the fluent $curr\_task(change_lane)$, and terminates the fluent $automation$ and $curr\_task(build\_sit\_aware)$.

\smallskip

\noindent\includegraphics[width=1.0\textwidth]{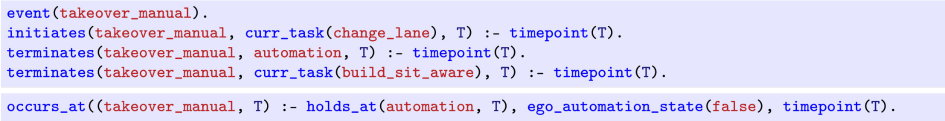}


\medskip 

\noindent \textbf{Generating the Driver's Scene Interpretation.}\quad Solving this Answer Set Program with the scene elements the driver is aware of results in a model of the scene based on the drivers subjective perception, which constitutes the interpretation model ($\mathcal{IM}$).

\smallskip

\noindent In particular this model includes the following elements:

\smallskip

\noindent\includegraphics[width=1.0\textwidth]{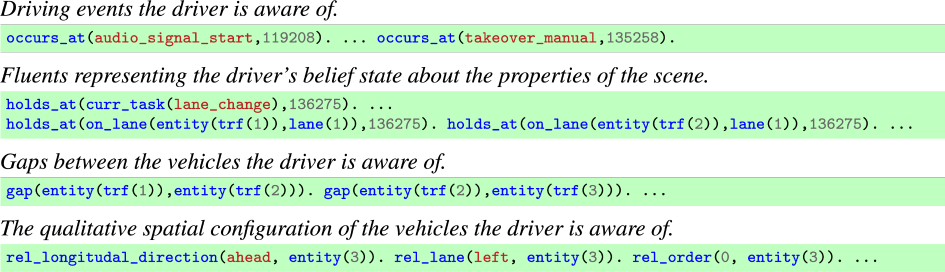}

%
%
%
%
%
%
%
%
%
%
%
%
%
%

\noindent This generated interpretation model ($\mathcal{IM}$) serves as input to the model comparison module to compare the drivers interpretation of the scene to the scene sensed by the SituSENSORS. Additionally these serve as abstractions needed for the projection step.

\medskip
\noindent \textbf{{\sffamily$\blacktriangleright$}} $~~$ {\sffamily\textbf{Projection.}} \quad
Within SituWare the projection step is used to explore possible future states by generating the set of events, which are possible in the current situation, and which are consistent with the task of the driver. For instance in our use-case the driver has to perform a lane change after taking over control from the automation system. To model this we define the lane change event within our domain characterisation and use the event calculus to generate possible lane change events the driver could perform for each time point. This set of possible events constitute the projection model ($\mathcal{PM}$) and can be used to identify scene elements the driver needs to be aware of in order to safely perform the given task.

\medskip 

\noindent \textbf{Event Anticipation.}\quad 
To anticipate scene events we define the events relevant for the driving task within the event calculus. 
For instance, the lane change event is defined based on the changes the event causes to the fluents and the constraints defining in which situations the event is possible.

\smallskip

\noindent\includegraphics[width=1.0\textwidth]{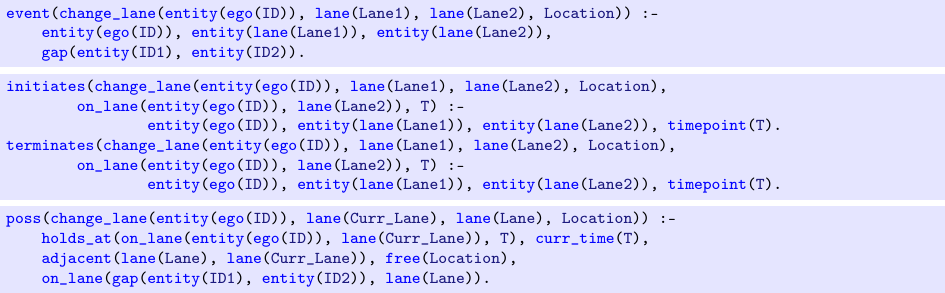}


\noindent In particular, a lane change is possible, when the lanes are adjacent to each other and there is a free location on the target lane. This free location may be either a gap between two vehicles, a free space behind or in front of a vehicle, or a completely empty lane.
For the projection of the scene we generate all events that initiate the goal of a particular task and that are possible in the current situation.

\noindent\includegraphics[width=1.0\textwidth]{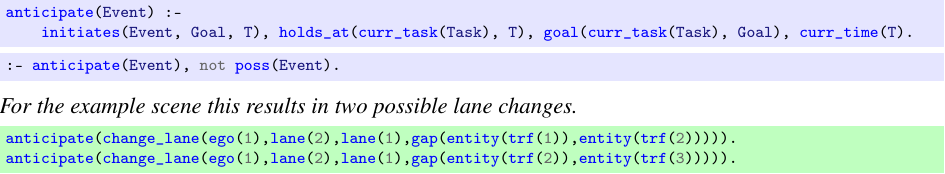}

%
%
%

\section{Results and Future Directions}
\label{sec:outlook}

We presented an online and real-time capable approach for modelling drivers' situation awareness in the context of takeover situations as they happen in semi-autonomous driving. The approach is based on declarative characterisations of driving dynamics and is implemented as a ROS module within Python and Answer Set Programming (ASP) as a part of \emph{SituSYS}, a modular system for measuring the drivers' situation awareness and guiding their attention based on the results.
We highlighted the application of the presented approach as part of the \emph{SituSYS} framework, which has been integrated in simulated as well as real-world driving. And we presented a case study concerned with situation awareness in take-over situations.

\medskip

\noindent \textbf{{\sffamily$\blacktriangleright$}} $~~$ {\sffamily\textbf{Application Results.}} \quad
Application of the interpretation and projection model in the context of the case study has shown that the model is capable of maintaining a representation of the scene and provide the necessary information for comparing the estimated scene representation of the driver with the sensed information, and to provide individual guidance using the specialized attention guidance devices implemented in the \emph{SituHMI}. The module works in real-time (approx. 30 Hz) with up to 20 vehicles plus the ego vehicle in the scene. However, to fully assess the effectiveness of the system it is required to perform a long term empirical study with the complete integrated system and measure the effect of \emph{SituSYS} on the driving performance in take-over situations. To conduct such a study is not in the scope of this paper and subject of ongoing and future research. 

\medskip

\noindent \textbf{{\sffamily$\blacktriangleright$}} $~~$ {\sffamily\textbf{Future Directions.}} \quad
The presented \emph{SituSYS} framework constitutes an ideal basis for applying logic based reasoning within simulated and real-world driving tasks. In this context future developments are aiming for more elaborate approaches for projecting scene dynamics, including the generation of possible and expected trajectories. 
Aside from this, an ongoing effort is the long term empirical assessment of the performance of \emph{SituSYS} with respect to the cognitive adequacy of the generated mental models within simulated as well as real-world driving.

\section*{Acknowledgments}

We acknowledge partial funding by the \emph{Federal Ministry for Economic Affairs and Climate Action} (BMWK) as part of the \emph{SituWare} project (reference no. 19A19011C and 19A19011F).

\bibliographystyle{eptcs}
\setlength{\bibsep}{2.8pt}


\begin{thebibliography}{10}
\providecommand{\bibitemdeclare}[2]{}
\providecommand{\surnamestart}{}
\providecommand{\surnameend}{}
\providecommand{\urlprefix}{Available at }
\providecommand{\url}[1]{\texttt{#1}}
\providecommand{\href}[2]{\texttt{#2}}
\providecommand{\urlalt}[2]{\href{#1}{#2}}
\providecommand{\doi}[1]{doi:\urlalt{https://doi.org/#1}{#1}}
\providecommand{\eprint}[1]{arXiv:\urlalt{https://arxiv.org/abs/#1}{#1}}
\providecommand{\bibinfo}[2]{#2}

\bibitemdeclare{inproceedings}{interactionPaper}
\bibitem{interactionPaper}
\bibinfo{author}{Mark \surnamestart Colley\surnameend}, \bibinfo{author}{Lukas
  \surnamestart Gruler\surnameend}, \bibinfo{author}{Marcel \surnamestart
  Woide\surnameend} \& \bibinfo{author}{Enrico \surnamestart Rukzio\surnameend}
  (\bibinfo{year}{2021}): \emph{\bibinfo{title}{Investigating the Design of
  Information Presentation in Take-Over Requests in Automated Vehicles}}.
\newblock In: {\slshape \bibinfo{booktitle}{Proceedings of the 23rd
  International Conference on Mobile Human-Computer Interaction}},
  \bibinfo{series}{MobileHCI '21}, \bibinfo{publisher}{Association for
  Computing Machinery}, \bibinfo{address}{New York, NY, USA},
  \doi{10.1145/3447526.3472025}.

\bibitemdeclare{inproceedings}{DBLP:conf/ecai/EiterK20}
\bibitem{DBLP:conf/ecai/EiterK20}
\bibinfo{author}{Thomas \surnamestart Eiter\surnameend} \&
  \bibinfo{author}{Rafael \surnamestart Kiesel\surnameend}
  (\bibinfo{year}{2020}): \emph{\bibinfo{title}{Weighted {LARS} for
  Quantitative Stream Reasoning}}.
\newblock In {\slshape \bibinfo{booktitle}{{ECAI} 2020 - 24th European Conference on
  Artificial Intelligence, 
  Santiago de Compostela,
  Spain,}}
  {\slshape \bibinfo{series}{Frontiers in Artificial Intelligence and
  Applications}} \bibinfo{volume}{325}, \bibinfo{publisher}{{IOS} Press},
  \doi{10.3233/FAIA200160}.


\bibitemdeclare{article}{endsley95}
\bibitem{endsley95}
\bibinfo{author}{Mica~R. \surnamestart Endsley\surnameend}
  (\bibinfo{year}{1995}): \emph{\bibinfo{title}{Toward a Theory of Situation
  Awareness in Dynamic Systems}}.
\newblock {\slshape \bibinfo{journal}{Human Factors}}
  \bibinfo{volume}{37}(\bibinfo{number}{1}), pp. \bibinfo{pages}{32--64},
  \doi{10.1518/001872095779049543}.

\bibitemdeclare{inproceedings}{Gebser2016-clingo5}
\bibitem{Gebser2016-clingo5}
\bibinfo{author}{Martin \surnamestart Gebser\surnameend},
  \bibinfo{author}{Roland \surnamestart Kaminski\surnameend},
  \bibinfo{author}{Benjamin \surnamestart Kaufmann\surnameend},
  \bibinfo{author}{Max \surnamestart Ostrowski\surnameend},
  \bibinfo{author}{Torsten \surnamestart Schaub\surnameend} \&
  \bibinfo{author}{Philipp \surnamestart Wanko\surnameend}
  (\bibinfo{year}{2016}): \emph{\bibinfo{title}{{Theory Solving Made Easy with
  Clingo 5}}}.
 \newblock In \bibinfo{booktitle}{Technical Communications of the 32nd International
  Conference on Logic Programming (ICLP 2016)}, {\slshape
  \bibinfo{series}{OpenAccess Series in Informatics
  (OASIcs)}}~\bibinfo{volume}{52}, \bibinfo{publisher}{SchlossDagstuhl--Leibniz-Zentrum fuer Informatik}, 
  \bibinfo{address}{Dagstuhl,
  Germany}, pp. \bibinfo{pages}{2:1--2:15}, \doi{10.4230/OASIcs.ICLP.2016.2}.
\newblock

\bibitemdeclare{book}{Gebser2012-ASP}
\bibitem{Gebser2012-ASP}
\bibinfo{author}{Martin \surnamestart Gebser\surnameend},
  \bibinfo{author}{Roland \surnamestart Kaminski\surnameend},
  \bibinfo{author}{Benjamin \surnamestart Kaufmann\surnameend} \&
  \bibinfo{author}{Torsten \surnamestart Schaub\surnameend}
  (\bibinfo{year}{2012}): \emph{\bibinfo{title}{Answer Set Solving in
  Practice}}.
\newblock \bibinfo{publisher}{Morgan \& Claypool Publishers},
  \doi{10.1007/978-3-031-01561-8}.

\bibitemdeclare{article}{Gebser2014-Clingo}
\bibitem{Gebser2014-Clingo}
\bibinfo{author}{Martin \surnamestart Gebser\surnameend},
  \bibinfo{author}{Roland \surnamestart Kaminski\surnameend},
  \bibinfo{author}{Benjamin \surnamestart Kaufmann\surnameend} \&
  \bibinfo{author}{Torsten \surnamestart Schaub\surnameend}
  (\bibinfo{year}{2014}): \emph{\bibinfo{title}{Clingo = {ASP} + Control:
  Preliminary Report}}.
\newblock {\slshape \bibinfo{journal}{CoRR}} \bibinfo{volume}{abs/1405.3694},
  \doi{10.48550/arXiv.1405.3694}.

\bibitemdeclare{inproceedings}{van2017priming}
\bibitem{van2017priming}
\bibinfo{author}{Remo~MA \surnamestart van~der Heiden\surnameend},
  \bibinfo{author}{Shamsi~T \surnamestart Iqbal\surnameend} \&
  \bibinfo{author}{Christian~P \surnamestart Janssen\surnameend}
  (\bibinfo{year}{2017}): \emph{\bibinfo{title}{Priming drivers before handover
  in semi-autonomous cars}}.
\newblock In: {\slshape \bibinfo{booktitle}{Proceedings of the 2017 CHI
  conference on human factors in computing systems}}, 
   \bibinfo{publisher}{ACM},
  \bibinfo{address}{New York, NY, USA}, pp.
  \bibinfo{pages}{392--404},
  \doi{10.1145/3025453.3025507}.

\bibitemdeclare{inproceedings}{hsiung2022}
\bibitem{hsiung2022}
\bibinfo{author}{Lei \surnamestart Hsiung\surnameend}, \bibinfo{author}{Yung-Ju
  \surnamestart Chang\surnameend}, \bibinfo{author}{Wei-Ko \surnamestart
  Li\surnameend}, \bibinfo{author}{Tsung-Yi \surnamestart Ho\surnameend} \&
  \bibinfo{author}{Shan-Hung \surnamestart Wu\surnameend}
  (\bibinfo{year}{2022}): \emph{\bibinfo{title}{A Lab-Based Investigation of
  Reaction Time and Reading Performance Using Different In-Vehicle Reading
  Interfaces during Self-Driving}}.
\newblock In: {\slshape \bibinfo{booktitle}{Proceedings of the 14th
  }}, 
  \bibinfo{series}{AutomotiveUI '22},
  \bibinfo{publisher}{ACM},
  \bibinfo{address}{New York, NY, USA}, p. \bibinfo{pages}{96 -- 107},
  \doi{10.1145/3543174.3545254}.

\bibitemdeclare{inproceedings}{DBLP:conf/iclp/KothawadeK0WG21}
\bibitem{DBLP:conf/iclp/KothawadeK0WG21}
\bibinfo{author}{Suraj \surnamestart Kothawade\surnameend},
  \bibinfo{author}{Vinaya \surnamestart Khandelwal\surnameend},
  \bibinfo{author}{Kinjal \surnamestart Basu\surnameend},
  \bibinfo{author}{Huaduo \surnamestart Wang\surnameend} \&
  \bibinfo{author}{Gopal \surnamestart Gupta\surnameend}
  (\bibinfo{year}{2021}): \emph{\bibinfo{title}{{AUTO-DISCERN:} Autonomous
  Driving Using Common Sense Reasoning}}.
  \newblock In: {\slshape \bibinfo{booktitle}{Proceedings of the International Conference on
  Logic Programming 2021 Workshops, 
  Porto, Portugal (virtual)
  }}, 
  {\slshape \bibinfo{series}{{CEUR} Workshop
  Proceedings}} \bibinfo{volume}{2970}, \bibinfo{publisher}{CEUR-WS.org}, \doi{10.48550/arXiv.2110.13606}.

\bibitemdeclare{inbook}{eventCalculus1989}
\bibitem{eventCalculus1989}
\bibinfo{author}{Robert \surnamestart Kowalski\surnameend} \&
  \bibinfo{author}{Marek \surnamestart Sergot\surnameend}
  (\bibinfo{year}{1989}): \emph{\bibinfo{title}{A Logic-Based Calculus of
  Events}}, pp. \bibinfo{pages}{23--51}.
\newblock \bibinfo{publisher}{Springer-Verlag}, \bibinfo{address}{Berlin,
  Heidelberg}, \doi{10.1007/978-3-642-83397-7_2}.

\bibitemdeclare{inproceedings}{krems2009}
\bibitem{krems2009}
\bibinfo{author}{Josef~F. \surnamestart Krems\surnameend} \&
  \bibinfo{author}{Martin R.~K. \surnamestart Baumann\surnameend}
  (\bibinfo{year}{2009}): \emph{\bibinfo{title}{Driving and Situation
  Awareness: A Cognitive Model of Memory-Update Processes}}.
\newblock In \bibinfo{editor}{Masaaki \surnamestart Kurosu\surnameend}, editor:
  {\slshape \bibinfo{booktitle}{Human Centered Design}},
  \bibinfo{publisher}{Springer Berlin Heidelberg}, pp.
  \bibinfo{pages}{986--994}, \doi{10.1007/978-3-642-02806-9_113}.

\bibitemdeclare{article}{kyriakidis2019-survey-human-factors}
\bibitem{kyriakidis2019-survey-human-factors}
\bibinfo{author}{M.~\surnamestart Kyriakidis\surnameend},
  \bibinfo{author}{J.~C.~F. \surnamestart de~Winter\surnameend},
  \bibinfo{author}{N.~\surnamestart Stanton\surnameend},
  \bibinfo{author}{T.~\surnamestart Bellet\surnameend},
  \bibinfo{author}{B.~\surnamestart van Arem\surnameend},
  \bibinfo{author}{K.~\surnamestart Brookhuis\surnameend},
  \bibinfo{author}{M.~H. \surnamestart Martens\surnameend},
  \bibinfo{author}{K.~\surnamestart Bengler\surnameend},
  \bibinfo{author}{J.~\surnamestart Andersson\surnameend},
  \bibinfo{author}{N.~\surnamestart Merat\surnameend},
  \bibinfo{author}{N.~\surnamestart Reed\surnameend},
  \bibinfo{author}{M.~\surnamestart Flament\surnameend},
  \bibinfo{author}{M.~\surnamestart Hagenzieker\surnameend} \&
  \bibinfo{author}{R.~\surnamestart Happee\surnameend} (\bibinfo{year}{2019}):
  \emph{\bibinfo{title}{A human factors perspective on automated driving}}.
\newblock {\slshape \bibinfo{journal}{Theoretical Issues in Ergonomics
  Science}} \bibinfo{volume}{20}(\bibinfo{number}{3}), pp.
  \bibinfo{pages}{223--249}, \doi{10.1080/1463922X.2017.1293187}.
  
\bibitemdeclare{article}{Le-Phuoc-Eiter-Le-Tuan-2021}
\bibitem{Le-Phuoc-Eiter-Le-Tuan-2021}
\bibinfo{author}{Danh \surnamestart Le-Phuoc\surnameend},
  \bibinfo{author}{Thomas \surnamestart Eiter\surnameend} \&
  \bibinfo{author}{Anh \surnamestart Le-Tuan\surnameend}
  (\bibinfo{year}{2021}): \emph{\bibinfo{title}{A Scalable Reasoning and
  Learning Approach for Neural-Symbolic Stream Fusion}}.
\newblock {\slshape \bibinfo{journal}{Proceedings of the AAAI Conference on
  Artificial Intelligence}} \bibinfo{volume}{35}(\bibinfo{number}{6}), pp.
  \bibinfo{pages}{4996--5005}, \doi{10.1609/aaai.v35i6.16633}.

\bibitemdeclare{inproceedings}{Ma2014-ASP-based-Epistemic-Event-Calculus}
\bibitem{Ma2014-ASP-based-Epistemic-Event-Calculus}
\bibinfo{author}{Jiefei \surnamestart Ma\surnameend}, \bibinfo{author}{Rob
  \surnamestart Miller\surnameend}, \bibinfo{author}{Leora \surnamestart
  Morgenstern\surnameend} \& \bibinfo{author}{Theodore \surnamestart
  Patkos\surnameend} (\bibinfo{year}{2014}): \emph{\bibinfo{title}{An Epistemic
  Event Calculus for ASP-based Reasoning About Knowledge of the Past, Present
  and Future}}.
\newblock In: {\slshape \bibinfo{booktitle}{{LPAR}: 19th International Conference on Logic
  for Programming, Artificial Intelligence and Reasoning}}, {\slshape
  \bibinfo{series}{EPiC Series in Computing}}~\bibinfo{volume}{26},
  \bibinfo{publisher}{EasyChair}, pp. \bibinfo{pages}{75--87},
  \doi{10.29007/zswj}.

\bibitemdeclare{inproceedings}{Miller2013-Epistemic-Event-Calculus}
\bibitem{Miller2013-Epistemic-Event-Calculus}
\bibinfo{author}{Rob \surnamestart Miller\surnameend}, \bibinfo{author}{Leora
  \surnamestart Morgenstern\surnameend} \& \bibinfo{author}{Theodore
  \surnamestart Patkos\surnameend} (\bibinfo{year}{2013}):
  \emph{\bibinfo{title}{Reasoning About Knowledge and Action in an Epistemic
  Event Calculus}}.
\newblock In: {\slshape \bibinfo{booktitle}{{COMMONSENSE 2013}}}.

\bibitemdeclare{inproceedings}{morgan2018}
\bibitem{morgan2018}
\bibinfo{author}{Phillip~L. \surnamestart Morgan\surnameend},
  \bibinfo{author}{Chris \surnamestart Alford\surnameend},
  \bibinfo{author}{Craig \surnamestart Williams\surnameend},
  \bibinfo{author}{Graham \surnamestart Parkhurst\surnameend} \&
  \bibinfo{author}{Tony \surnamestart Pipe\surnameend} (\bibinfo{year}{2018}):
  \emph{\bibinfo{title}{Manual Takeover and Handover of a Simulated Fully
  Autonomous Vehicle Within Urban and Extra-Urban Settings}}.
\newblock In 
  {\slshape \bibinfo{booktitle}{Advances in Human Aspects of
  Transportation}}, \bibinfo{publisher}{Springer International Publishing},
  \bibinfo{address}{Cham}, pp. \bibinfo{pages}{760--771},
  \doi{10.1007/978-3-319-60441-1_73}.

\bibitemdeclare{inproceedings}{iccm2017}
\bibitem{iccm2017}
\bibinfo{author}{Jan-Patrick \surnamestart Osterloh\surnameend},
  \bibinfo{author}{Jochem~W. \surnamestart Rieger\surnameend} \&
  \bibinfo{author}{Andreas \surnamestart L{\"u}dtke\surnameend}
  (\bibinfo{year}{2017}): \emph{\bibinfo{title}{Modelling Workload of a Virtual
  Driver}}.
\newblock In: {\slshape \bibinfo{booktitle}{Proceedings of the 15th
  International Conference on Cognitive Modeling}}.

\bibitemdeclare{article}{rasouli2020}
\bibitem{rasouli2020}
\bibinfo{author}{Amir \surnamestart Rasouli\surnameend} \&
  \bibinfo{author}{John~K. \surnamestart Tsotsos\surnameend}
  (\bibinfo{year}{2020}): \emph{\bibinfo{title}{Autonomous Vehicles That
  Interact With Pedestrians: A Survey of Theory and Practice}}.
\newblock {\slshape \bibinfo{journal}{IEEE Transactions on Intelligent
  Transportation Systems}} \bibinfo{volume}{21}(\bibinfo{number}{3}), pp.
  \bibinfo{pages}{900--918}, \doi{10.1109/TITS.2019.2901817}.

\bibitemdeclare{article}{rehman2019}
\bibitem{rehman2019}
\bibinfo{author}{Umair \surnamestart Rehman\surnameend}, \bibinfo{author}{Shi
  \surnamestart Cao\surnameend} \& \bibinfo{author}{Carolyn \surnamestart
  MacGregor\surnameend} (\bibinfo{year}{2019}): \emph{\bibinfo{title}{Using an
  Integrated Cognitive Architecture to Model the Effect of Environmental
  Complexity on Drivers' Situation Awareness}}.
\newblock {\slshape \bibinfo{journal}{Proceedings of the Human Factors and
  Ergonomics Society Annual Meeting}}
  \bibinfo{volume}{63}(\bibinfo{number}{1}), pp. \bibinfo{pages}{812--816},
  \doi{10.1177/1071181319631313}.


\bibitemdeclare{inproceedings}{roedel2014}
\bibitem{roedel2014}
\bibinfo{author}{Christina \surnamestart R\"{o}del\surnameend},
  \bibinfo{author}{Susanne \surnamestart Stadler\surnameend},
  \bibinfo{author}{Alexander \surnamestart Meschtscherjakov\surnameend} \&
  \bibinfo{author}{Manfred \surnamestart Tscheligi\surnameend}
  (\bibinfo{year}{2014}): \emph{\bibinfo{title}{Towards Autonomous Cars: The
  Effect of Autonomy Levels on Acceptance and User Experience}}.
\newblock In: {\slshape \bibinfo{booktitle}{Proceedings of the 6th
  }}, \bibinfo{series}{AutomotiveUI '14},
  \bibinfo{publisher}{ACM},
  \bibinfo{address}{New York, NY, USA}, p. \bibinfo{pages}{1 -- 8},
  \doi{10.1145/2667317.2667330}.


\bibitemdeclare{inproceedings}{Suchan-IJCAI2019}
\bibitem{Suchan-IJCAI2019}
\bibinfo{author}{Jakob \surnamestart Suchan\surnameend}, \bibinfo{author}{Mehul
  \surnamestart Bhatt\surnameend} \& \bibinfo{author}{Srikrishna \surnamestart
  Varadarajan\surnameend} (\bibinfo{year}{2019}): \emph{\bibinfo{title}{Out of
  Sight But Not Out of Mind: An Answer Set Programming Based Online Abduction
  Framework for Visual Sensemaking in Autonomous Driving}}.
\newblock In 
  {\slshape \bibinfo{booktitle}{Proceedings of the 28th International
  Joint Conference on Artificial Intelligence, {IJCAI} 2019, Macao, China}},
  \bibinfo{publisher}{ijcai.org}, pp. \bibinfo{pages}{1879--1885},
  \doi{10.24963/ijcai.2019/260}.

\bibitemdeclare{article}{DBLP:journals/ai/SuchanBV21}
\bibitem{DBLP:journals/ai/SuchanBV21}
\bibinfo{author}{Jakob \surnamestart Suchan\surnameend}, \bibinfo{author}{Mehul
  \surnamestart Bhatt\surnameend} \& \bibinfo{author}{Srikrishna \surnamestart
  Varadarajan\surnameend} (\bibinfo{year}{2021}):
  \emph{\bibinfo{title}{Commonsense visual sensemaking for autonomous driving -
  On generalised neurosymbolic online abduction integrating vision and
  semantics}}.
\newblock {\slshape \bibinfo{journal}{Artificial Intelligence}} \bibinfo{volume}{299},
  p. \bibinfo{pages}{103522}, \doi{10.1016/j.artint.2021.103522}.

\bibitemdeclare{inproceedings}{Suchan18}
\bibitem{Suchan18}
\bibinfo{author}{Jakob \surnamestart Suchan\surnameend}, \bibinfo{author}{Mehul
  \surnamestart Bhatt\surnameend}, \bibinfo{author}{Przemyslaw~Andrzej
  \surnamestart Walega\surnameend} \& \bibinfo{author}{Carl \surnamestart
  Schultz\surnameend} (\bibinfo{year}{2018}): \emph{\bibinfo{title}{Visual
  Explanation by High-Level Abduction: On Answer-Set Programming Driven
  Reasoning About Moving Objects}}.
 \newblock In {\slshape \bibinfo{booktitle}{Proceedings of the 32nd {AAAI}
  Conference on Artificial Intelligence, (AAAI-18), New Orleans, Louisiana,
  USA}}, \bibinfo{publisher}{{AAAI} Press}, \doi{10.1609/aaai.v32i1.11569}.

\bibitemdeclare{article}{tsilionis-koutroumanis-nikitopoulos-doulkeridis-artikis-2019}
\bibitem{tsilionis-koutroumanis-nikitopoulos-doulkeridis-artikis-2019}
\bibinfo{author}{Efthimis \surnamestart Tsilionis\surnameend},
  \bibinfo{author}{Nikolaos \surnamestart Koutroumanis\surnameend},
  \bibinfo{author}{Panagiotis \surnamestart Nikitopoulos\surnameend},
  \bibinfo{author}{Christos \surnamestart Doulkeridis\surnameend} \&
  \bibinfo{author}{Alexander \surnamestart Artikis\surnameend}
  (\bibinfo{year}{2019}): \emph{\bibinfo{title}{Online Event Recognition from
  Moving Vehicles: Application Paper}}.
\newblock {\slshape \bibinfo{journal}{Theory and Practice of Logic
  Programming}} \bibinfo{volume}{19}(\bibinfo{number}{5-6}),
  \doi{10.1017/S147106841900022X}.

\bibitemdeclare{}{OpenDrive}
\bibitem{OpenDrive}
\bibinfo{author}{ASAM \surnamestart e.~V.\surnameend} (\bibinfo{year}{2021}):
  \emph{\bibinfo{title}{OpenDRIVE Format Specification}}.
\newblock \urlprefix\url{https://www.asam.net/standards/detail/opendrive/}.

\bibitemdeclare{inproceedings}{zimmermann2017first}
\bibitem{zimmermann2017first}
\bibinfo{author}{Raphael \surnamestart Zimmermann\surnameend} \&
  \bibinfo{author}{Reto \surnamestart Wettach\surnameend}
  (\bibinfo{year}{2017}): \emph{\bibinfo{title}{First step into visceral
  interaction with autonomous vehicles}}.
\newblock In: {\slshape \bibinfo{booktitle}{Proceedings of the 9th
  International Conference on Automotive User Interfaces and Interactive
  Vehicular Applications}}, pp. \bibinfo{pages}{58--64},
  \doi{10.1145/3122986.3122988}.

\end{thebibliography}

\end{document}